\documentclass[letterpaper]{article} 
\usepackage{aaai25}  
\usepackage{times}  
\usepackage{helvet}  
\usepackage{courier}  
\usepackage[hyphens]{url}  
\usepackage{graphicx} 
\usepackage{natbib}  
\usepackage{caption} 
\usepackage{algorithm}
\usepackage{algorithmic}
\usepackage{bbding}
\usepackage{booktabs} 

\usepackage{newfloat}
\usepackage{listings}
\usepackage{multirow}
\DeclareCaptionStyle{ruled}{labelfont=normalfont,labelsep=colon,strut=off} 
\floatstyle{ruled}
\newfloat{listing}{tb}{lst}{}
\floatname{listing}{Listing}
\usepackage{algorithm}
\usepackage{algorithmic}
\usepackage{amsmath, amsthm, amssymb} 
\usepackage{mathrsfs}
\usepackage{newfloat}
\usepackage{listings}
\usepackage{float}

\title{DiT4Edit: Diffusion Transformer for Image Editing}

\author{
   \\
    Kunyu Feng\textsuperscript{\rm 1}\equalcontrib,
    Yue Ma\textsuperscript{\rm 2}\equalcontrib,
    Bingyuan Wang\textsuperscript{\rm 3}\equalcontrib,
    Chenyang Qi\textsuperscript{\rm 2},
    Haozhe Chen\textsuperscript{\rm 1},\\
    Qifeng Chen\textsuperscript{\rm 2}\footnote{Corresponding Author.},
    Zeyu Wang\textsuperscript{\rm 3}\textsuperscript{\dag}
}

\affiliations{
    \textsuperscript{\rm 1}Peking University\\
    \textsuperscript{\rm 2}The Hong Kong University of Science and Technology\\
    \textsuperscript{\rm 3}Hong Kong University of Science and Technology (Guangzhou)

}

\usepackage{bibentry}

\begin{document}
\maketitle
\begin{abstract}
Despite recent advances in UNet-based image editing, methods for shape-aware object editing in high-resolution images are still lacking. Compared to UNet, Diffusion Transformers (DiT) demonstrate superior capabilities to effectively capture the long-range dependencies among patches, leading to higher-quality image generation. In this paper, we propose DiT4Edit\footnote{Project page: https://github.com/fkyyyy/DiT4Edit}, the first Diffusion Transformer-based image editing framework. Specifically, DiT4Edit uses the DPM-Solver inversion algorithm to obtain the inverted latents, reducing the number of steps compared to the DDIM inversion algorithm commonly used in UNet-based frameworks. Additionally, we design unified attention control and patches merging, tailored for transformer computation streams. This integration allows our framework to generate higher-quality edited images faster. Our design leverages the advantages of DiT, enabling it to surpass UNet structures in image editing, especially in high-resolution and arbitrary-size images. 
Extensive experiments demonstrate the strong performance of DiT4Edit across various editing scenarios, highlighting the potential of Diffusion Transformers in supporting image editing.
\end{abstract}

\section{Introduction}

Recent advances in diffusion models have witnessed impressive progress in text-driven visual generation. The development of these text-to-image (T2I) models, e.g., Stable Diffusion (SD)~\cite{rombach2022high}, DALL$\cdot$E 3~\cite{betker2023improving}, and PixArt~\cite{chen2023pixart}, has led to significant impacts on numerous downstream applications~\cite{ma2024followyourpose}~\cite{ma2024followyourclick}~\cite{wang2024cove}, with image editing as one of the most challenging tasks. 
Given a synthetic or real input image, image editing algorithms aim to add, remove, or replace entire objects or object attributes according to the user's intent.

A primary challenge in text-driven image editing is maintaining the consistency between the source and target images. Earlier approaches~\cite{choi2021ilvr}~\cite{Kawar_2023_CVPR}~\cite{Zhang_2023_CVPR} often relied on fine-tuning diffusion models to address this issue. However, these methods typically require considerable time and computation resources, which limits their practical applicability. Recent approaches often utilize DDIM~\cite{DBLP:journals/corr/abs-2010-02502} inversion to obtain latent maps and then control the attention mechanism in diffusion models for real image editing~\cite{Mokady_2023_CVPR}~\cite{cao2023masactrl}. However, the consistency of the edited images depends heavily on the invertibility of the DDIM inversion process. Although some efforts have focused on optimizing this inversion~\cite{ju2024pnp}~\cite{Dong_2023_ICCV} for better results, the editing framework still relies on too many timesteps (e.g., 50 steps).

In addition, current research on image editing tasks mainly uses the UNet-based diffusion model structure~\cite{rombach2022high}, making the final editing results heavily bounded by the generative capacity of UNet.
Although the attention mechanism in UNet is also derived from the transformer, DiT~\cite{Peebles_2023_ICCV} based on pure transformers offers a global attention calculation between patches, allowing them to capture broader and more detailed features compared to the UNet with convolution blocks, leading to higher-quality images. In addition, evidence from DiT demonstrates that transformer-based diffusion models offer better scalability and outperform UNet-based models in large-scale experiments.


To address these challenges, we explore image editing tasks using the diffusion transformer architecture and provide a valuable empirical baseline for future research. First, we aim to leverage solvers that require fewer inversion steps to reduce our inference time while maintaining the image quality of the results.
Specifically, we employ an inversion algorithm based on a high-order DPM-Solver~\cite{lu2023dpmsolverfastsolverguided} to obtain better latent maps with fewer timesteps. We then implement a unified attention control scheme for text-guided image editing while preserving background details. Third, to mitigate the increased computational complexity of transformers compared to UNet, we use patches merging to accelerate computation. By integrating these key components, we introduce DiT4Edit, the first diffusion transformer-based editing framework to our knowledge. Experiments demonstrate that our framework achieves superior editing results with fewer inference steps and offers distinct advantages over traditional UNet-based methods.

\twocolumn[{
\renewcommand\twocolumn[1][]{#1}%
\begin{center}
    \vspace{-1\baselineskip}
    \includegraphics[width=0.96\textwidth]{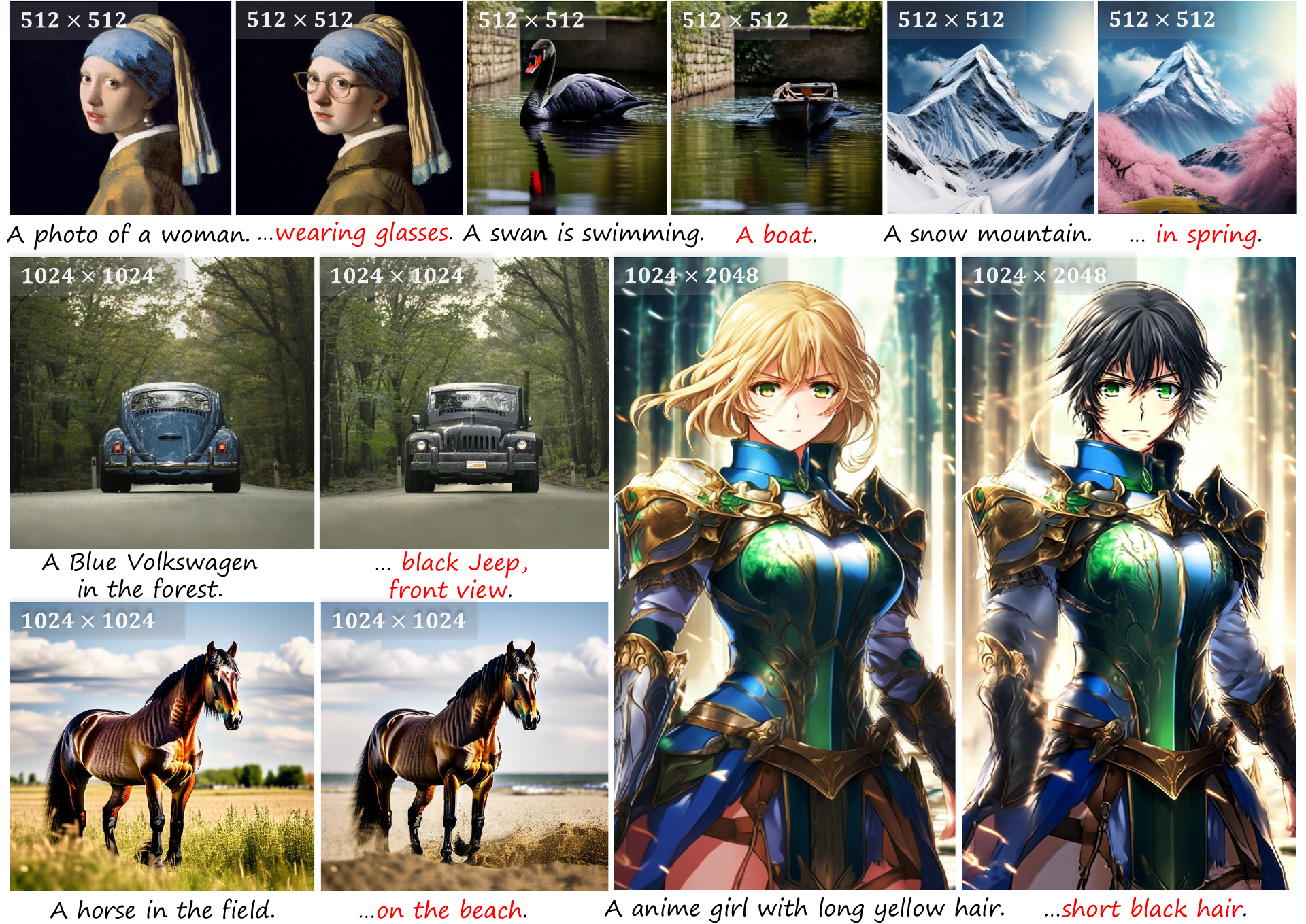}
     \vspace{-0.05cm}
    \captionof{figure}{
    \textbf{Visual results of DiT4Edit.} Our method is the first DiT-based image editing framework, which is capable of handling images of various sizes: from small ($512\times512$) to large ($1024\times1024$), and even arbitrary dimensions (up to $1024\times2048$).
    }
    \label{fig:FigureVisualresults}
    
\end{center}
}]
In summary, our contributions are as follows:
\begin{itemize}
    \item Based on the advantages of transformer-based diffusion models in image editing, we introduce DiT4Edit, the first tuning-free image editing framework using Diffusion Transformers (DiT). 
    \item To adapt to the computing mechanism of transformer-based denoising, we first propose a unified attention control mechanism to achieve image editing. Then, we introduce the DPM-Solver inversion and patches merging strategy to reduce inference time.
    \item Extensive qualitative and quantitative results demonstrate the superior performance of DiT4Edit in object editing, style editing, and shape-aware editing for various image sizes, including $512 \times 512$, $1024 \times 1024$, $1024 \times 2048$. 
\end{itemize}

\section{Related Work}

\subsection{Text-to-Image Generation}
Since Dosovitskiy et al.~\cite{dosovitskiy2020image} introduced the Visual Transformer (ViT) and highlighted the potential of transformer architectures for image tasks, numerous transformer-based visual applications have been developed, including high-resolution image synthesis~\cite{esser2021taming}. 
Prior to the advent of diffusion models, researchers predominantly relied on a generative adversarial network (GAN) for image synthesis~\cite{goodfellow2014generative}. Zhang et al. ~\cite{zhang2018photographic} developed a single-stream generator capable of producing high-resolution images, while Liang et al. ~\cite{liang2020cpgan} enhanced the performance of text-to-image synthesis by incorporating Memory-Attended Text Encoder and Object-Aware Image Encoder. Subsequently, the Denoising Diffusion Probabilistic Models (DDPMs) introduced by Ho et al.~\cite{ho2020denoising} marked significant leaps forward, achieving breakthroughs in image quality, controllability, and diversity.
The designs and applications of diffusion-based methods can be classified by tasks such as controllable generation, stylization, and quality improvement. Dhariwal et al.~\cite{dhariwal2021diffusion} introduces a classifier guidance method to improve the generation quality of diffusion models, while Yang et al.~\cite{yang2024improving} uses the CLIP latents to produce realistic images closely aligned with human expectations in text-to-image generation tasks. In recent text-to-image generation tasks, ControlNet~\cite{zhang2023adding} allows for the integration of user-specified conditional information into the image generation process. Meanwhile, ScaleCrafter~\cite{he2023scaleCrafter} addresses the issue of limited perception in convolutional layers during the diffusion model generation process, enabling the production of higher resolution and higher quality images. Moreover, these T2I models have also driven the development of a series of video generation and editing applications~\cite{ma2022visual,ma2023magicstick,ma2022simvtp,ma2024followyouremoji,chen2024follow}.

\subsection{Interactive Image Editing}
Image editing encompasses scenarios like iterative generation, collaborative creation, and image inpainting. Research has focused on decoupling high-level concepts and low-level styles within deep latent structures to improve diffusion-based models' performance in tasks such as content editing (detail control)~\cite{kawar2023imagic}, style transfer~\cite{brack2022stable}, and textual inversion~\cite{gal2022image}.
Compared to other generative models, diffusion models offer enhanced controllability during the image generation process, allowing for precise manipulation of image attributes \cite{choi2021ilvr} \cite{zhang2023adding}. These advantages enable diffusion models to achieve outstanding performance in image editing tasks.
Hertz et al.~\cite{hertz2022prompt} introduced a framework for image editing through textual prompts, which transforms the original image into the target image by modifying, adding, and adjusting the weights of the cross-attention map. Methods like InstructPix2Pix \cite{brooks2023instructpix2pix} and Custom Diffusion \cite{kumari2023multi} employ user-guided approaches to achieve image editing. These techniques allow for modifications by inputting various types of guiding prompts, allowing diffusion models to swiftly adapt to new concepts. Parmar et al.~\cite{parmar2023zero} employed pix2pix-zero to address the challenge of preserving the original structure while incorporating user-specified changes during image editing.
Although there have been significant advancements in image editing using diffusion models, these existing attempts at image editing are still bounded by the pretrained generative power of a UNet. 
Compared with UNet-based diffusion models, DiT is more scalable and has more succinct architectures, while DiT's application in image editing is still under-explored.

\begin{figure*}[t]
    \centering
    \includegraphics[width=.98\textwidth]{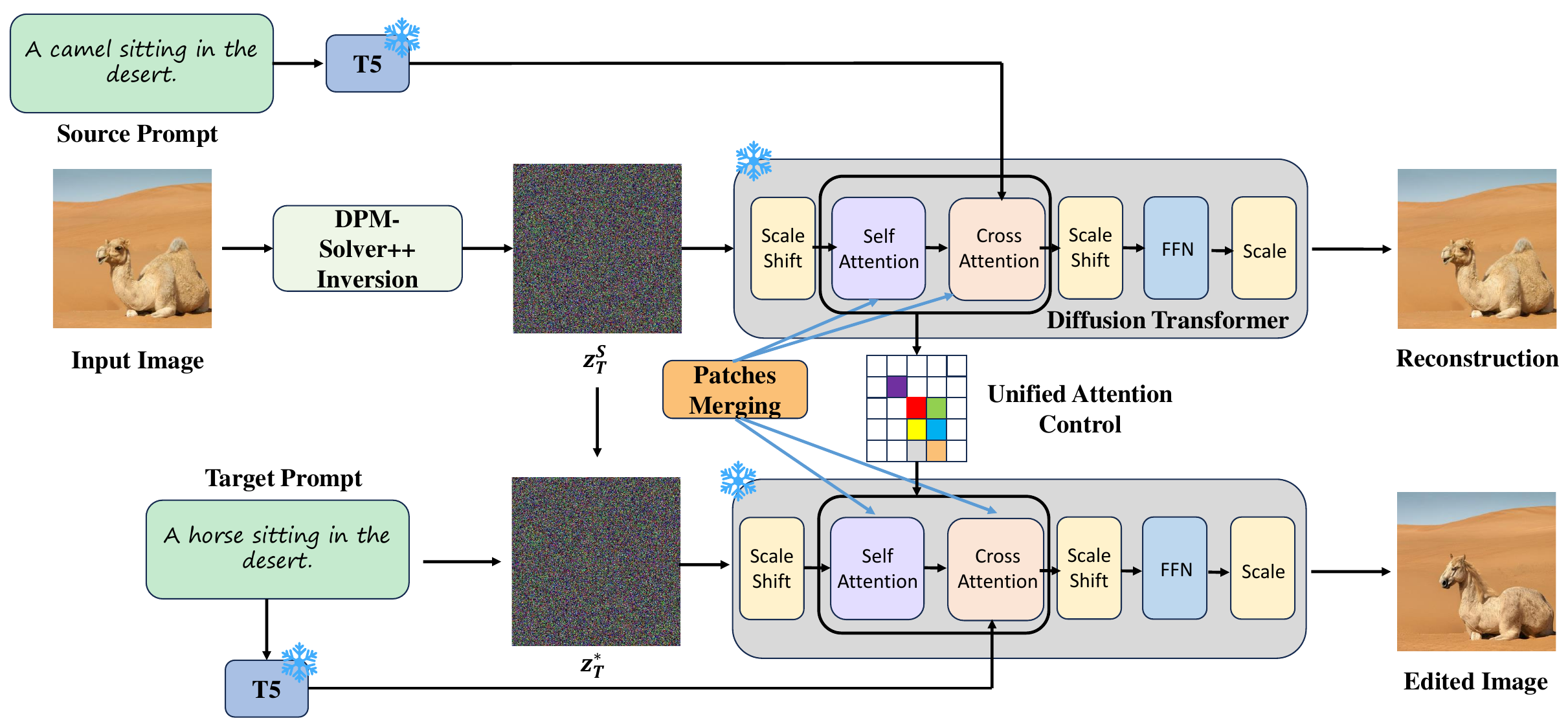}
    \vspace{-0.2cm}
    \caption{Overview of the DiT4Edit framework. During the image editing process, our inversion algorithm generates high-quality latent maps, and the final edited image is achieved through unified attention control.
    } 
    \label{fig:high-res}
\end{figure*}

\section{Methodology}
Our proposed framework aims to achieve high-quality image editing for various sizes based on a diffusion transformer. Our method is the first editing strategy based on a pre-trained text-to-image transformer-based diffusion model, e.g., PIXART-$\alpha$ ~\cite{chen2023pixart}. With our approach, users can achieve better editing results compared to existing UNet-based methods by providing a target prompt. 
In this section, we first introduce the latent diffusion models and DPM inversion. Then we illustrate the superiority of transformer-based denoising in image editing tasks. Finally, we discuss the implementation details of our editing framework.
\subsection{Preliminaries: Latent Diffusion Models}
The Latent Diffusion Model (LDM)~\cite{rombach2022high} proposes an image generation method with a denoising process within a latent space $\mathcal{Z}$. In particular, it uses a pre-trained image encoder $\mathcal{E}$ to encode the input image $x$ into low-resolution latents $z=\mathcal{E}(x)$. During training, the model optimizes a denoising UNet $\epsilon_\theta$ by removing artificial noise, conditioned on both text prompt embedding $y$ and current image sample $z_t$, which is a noisy sample of $z_0$ at step $t\in[0, T]$:
\begin{equation}
\min _\theta E_{z_0, \epsilon \sim N(0, I), t}\left\|\epsilon-\epsilon_\theta\left(z_t, t, y\right)\right\|_2^2,
\label{equation: denosing}
\end{equation}
After training, it is capable of converting random noise $\epsilon$ to an image sample $z$ by the learned denoising process.

\textbf{DPM-Solver Sampling.}
During the inversion stage in Diffusion probabilistic models (DPMs), a clean image $x{_0}$ is gradually added 
 with Gaussian noise and turned into a noisy sample ${x_t}$:
\begin{equation}
q(\boldsymbol x_t | \boldsymbol x_0) = \mathcal{N}(\boldsymbol x_t | \alpha_t\boldsymbol x_0, \sigma_t^2\boldsymbol I),
\label{equation: inversion}
\end{equation}
where $\frac{\alpha_t^2}{\sigma_t^2}$ is the signal-to-noise ratio (SNR), which is a strictly decreasing function of $t$ ~\cite{lu2023dpmsolverfastsolverguided}. Through solving the diffusion ODE, the DPM sampling can be faster than other methods:
\begin{equation}
\frac{d\boldsymbol x_t}{dt} = (f(t) + \frac{g^2(t)}{2\sigma_t^2})\boldsymbol x_t - \frac{\alpha_tg^2(t)}{2\sigma_t^2}\boldsymbol x_\theta(\boldsymbol x_t, t),
\label{equation: DPM Sampling by ODE}
\end{equation}
where $\boldsymbol x_T\sim\mathcal{N}(\boldsymbol 0, \widetilde \alpha^2,\boldsymbol I)$, and $f(t)=\frac{\mathrm{dlog}\alpha_t}{\mathrm{d}t}$, $g(t)=\frac{\mathrm{d}\alpha_t^2}{\mathrm{d}t} - 2\frac{\mathrm{dlog}\alpha_t}{\mathrm{d}t}\alpha_t^2$~\cite{lu2023dpmsolverfastsolverguided}. It is shown in the previous works ~\cite{lu2022dpm} ~\cite{zhang2022fast} that ODE solver using the exponential integrator exhibits faster convergence compared to traditional solvers during solving the Eq.~\ref{equation: DPM Sampling by ODE}. By setting the value of $\boldsymbol x_s$, the solution $\boldsymbol x_t$ of Eq.~\ref{equation: DPM Sampling by ODE} can be calculated by:
\begin{equation}
\boldsymbol x_t = \frac{\alpha_t}{\alpha_s}\boldsymbol x_s - \alpha_t\int_{\lambda_s}^{\lambda_t} e^{-\lambda}\boldsymbol x_\theta(\hat{\boldsymbol x_\lambda}, \lambda)\mathrm{d}\lambda,
\label{equation: The solution of Eq3}
\end{equation}
where the $\lambda_t=\log(\alpha_t/\sigma_t)$ is a decreasing function of t with the inversion function $t_\lambda(\cdot)$, and recent research demonstrates that the DPM-Solver can sample the realistic images in 10--20 steps.

\subsection{Diffusion Model Architecture}\textbf{PIXART-$\alpha$.}
Compared to the UNet structure, Diffusion Transformers (DiT)~\cite{Peebles_2023_ICCV} exhibits superior scaling properties, generating images of higher quality and demonstrating better performance. 

PIXART-$\alpha$ is a Transformer-based text-to-image (T2I) diffusion model that consists of three main components: Cross-Attention layer, AdaLN-single, and Re-parameterization.~\cite{chen2023pixart} Researchers have trained this T2I diffusion model with three sophisticated designs: decomposition training strategies, efficient T2I transformer and high-informative data. Many experimental results demonstrate that PIXART-$\alpha$ performs better in image quality, artistry, and semantic control. Compared to advanced T2I SOTA models, PIXART-$\alpha$ has faster training speed, lower inference cost, and superior comprehensive performance. In this paper, we use PIXART-$\alpha$ as the baseline for our proposed image editing method. 

\textbf{The reason for using a transformer as the denoising model.} Compared to the UNet structure, the transformer incorporates a global attention mechanism, allowing the model to focus on a broader range within the image. This enhanced scalability enables transformers to generate high-quality images at large sizes (e.g., greater than $512\times512$), and even at arbitrary sizes. The editing results of our DiT-based editing framework for large-sized images are demonstrated in Figures 1 and 2. These represent editing tasks not previously addressed by UNet-based frameworks. Therefore, we adopted a transformer-based denoising model for our editing framework, leveraging the transformer's capabilities to tackle these more complex editing challenges.
\begin{figure*}
    \centering
    \includegraphics[width=\textwidth]{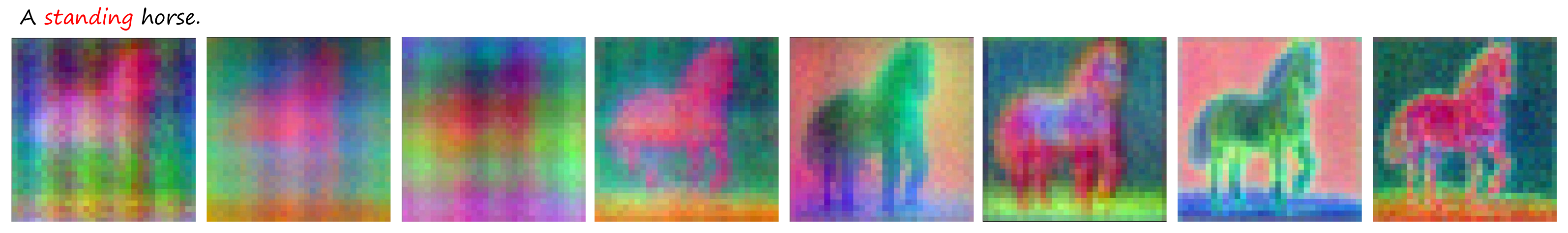}
    \caption{Visualization of the \textit{Query} features in the self-attention layers of the PixArt-$\alpha$. Features in the deeper layers (right side) are observed to capture the semantic layout more effectively than those in the shallow layers (left side).
    }
    \label{fig:attn_map}
\end{figure*}
\subsection{Diffusion Transformer-based Image Editing}
In this section, we introduce the components of our proposed DiT4Edit. As shown in Figure~\ref{fig:high-res}, based on a pretrained diffusion transformer, the pipeline of our image editing framework is as follows.

\textbf{DPM-Solver inversion.} As we discussed earlier, using a high-order DPM-Solver (e.g., DPM-Solver++),  can effectively improve the sampling speed. To approximate the integral term $\int_{\lambda_s}^{\lambda_t} e^{-\lambda}\boldsymbol x_\theta\mathrm{d}\lambda$\, in equation Eq.~\ref{equation: The solution of Eq3}, given the $x_{t_{i-1}}$ at time $t_{t_{i-1}}$, using the Taylor expansion at $\lambda_{t_{i-1}}$, and the DPM-Solver++ can obtain a exact solution value at time $t_i$:
\begin{equation}
\begin{split}
\boldsymbol x_{t_i} &=
 \frac{\sigma_{t_i}}{\sigma_{t_{i-1}}}\boldsymbol x_{t_{i-1}} + \sigma_{t_i}
 \sum_{n=0}^{k-1} 
 \underbrace{\boldsymbol x_{\theta}^{(n)}(\boldsymbol x_{\lambda_{t_{i-1}}}, \lambda_{t_{i-1}})}_{\text{estimated}}
 \\
 &\int_{\lambda_{t_{i-1}}}^{\lambda_{t_{i}}} 
 \underbrace{e^\lambda\frac{(\lambda-\lambda_{t_{i-1}})^n}{n!}\mathrm{d}\lambda}_{\text{analtically computed}} + \underbrace{\mathcal{O}(h_{i}^{k+1})}_{\text{omitted}},   
 \end{split}
\label{equation: DPM-Solver++}
\end{equation}
Especially when $k=1$, the Eq.~\ref{equation: DPM-Solver++} is equivalent to DDIM sampler~\cite{song2021denoising} as follows:
\begin{equation}
\boldsymbol x_{t_i} =
 \frac{\sigma_{t_i}}{\sigma_{t_{i-1}}}\boldsymbol x_{t_{i-1}} - \alpha_{t_{i}}(e^{-h_{i}}-1)\boldsymbol x_\theta(\boldsymbol x_{t_{i-1}},t_{i-1}),
\label{equation: DDIM Sampler}
\end{equation}
In practical applications, it is common to set $k = 2$, enabling a rapid inference and minimizing discretization errors. This DPM-Solver named DPM-Solver++ (2M)~\cite{lu2023dpmsolverfastsolverguided}:
\begin{equation}
\begin{split}
\boldsymbol x_{t_i} &=
 \frac{\sigma_{t_i}}{\sigma_{t_{i-1}}}\boldsymbol x_{t_{i-1}}- \alpha_{t_{i}}(e^{-h_{i}}-1)\,\cdot[
 \\
 &[(1 + \frac{1}{2r_i})\boldsymbol x_\theta(x_{t_{i-1}}, t_{i-1}) - \frac{1}{2r_i}\boldsymbol x_\theta(\boldsymbol x_{t_{i-2}}, t_{i-2})],
 \end{split}
\label{equation: DPM-Solver++(2M)}
\end{equation}
where 2M means this solver is a second-order multistep solver.

However, during the inversion stage for the high-order samplers such as DPM-Solver++(2M), to obtain the inversion result $\boldsymbol x_{t_i}$ in current timestep $t_i$, we need to approximate the values in prior timesteps like \{$t_{i-2}, t_{i-3},...$\} for the estimated and the analytically computed terms in Eq.~\ref{equation: DPM-Solver++}:
\begin{equation}
\sigma_{t_i}
 \sum_{n=0}^{k-1} 
 \boldsymbol x_{\theta}^{(n)}(\boldsymbol x_{\lambda_{t_{i-1}}}, \lambda_{t_{i-1}})\int_{\lambda_{t_{i-1}}}^{\lambda_{t_{i}}} 
 e^\lambda\frac{(\lambda-\lambda_{t_{i-1}})^n}{n!}\mathrm{d}\lambda,
 \label{equation: estimated terms in DPM-solver inversion}
\end{equation}
A recent work~\cite{Hong_2024_CVPR} introduced a strategy via the backward Euler method to get the high-order term approximation in Eq.~\ref{equation: estimated terms in DPM-solver inversion} as follows:
\begin{equation}
\boldsymbol d_{i}^\prime = \boldsymbol z_\theta(\hat{z}_{t_{i-1}}, t_{i-1}) + \frac{\boldsymbol z_\theta(\boldsymbol{\hat{y}}_{t_{i-1}}, t_{i-1}) - \boldsymbol z_\theta(\boldsymbol{\hat{y}}_{t_{i-2}}, t_{i-2})}{2r_{i}},
\label{equation: estimated the high-order terms in DPM-solver inversion}
\end{equation}
where $\boldsymbol z_\theta$ is the denosing model, \{$\boldsymbol{\hat{y}}_{t_{i-1}}, \boldsymbol{\hat{y}}_{t_{i-2}}, ...$\} is a set of value calculated by $\hat{x}_{t_{i}}$ through the DDIM inversion to estimate the ($\hat{x}_{t_{i-1}}, \hat{x}_{t_{i-2}}$) in Eq.~\ref{equation: estimated terms in DPM-solver inversion}, and $r_{i} = \frac{\lambda_{t_{i-1}}-\lambda_{t_{i-2}}}{\lambda_{t_i}-\lambda_{t_{i-1}}}$. Then we can get a inversion latent $\hat{z}_{t_{i-1}}$ in current timestep by:

\begin{equation}
\boldsymbol{\hat{z}}_{t_{i-1}} = \boldsymbol{\hat{z}_{t_{i-1}}} - \rho(\boldsymbol z_{t_i}^\prime - \boldsymbol{\hat{z}}_{t_i}),
\label{equation: perfect inversion latent}
\end{equation}
where $\boldsymbol z_{t_i}^\prime = \frac{\sigma_{t_{i}}}{\sigma_{t_{i-1}}}\boldsymbol{\hat{z}}_{t_{i-1}}-\alpha_{t_{i}}(e^{-h_{i}}-1)\boldsymbol d_{i}^\prime$. In DiT4Edit, we utilize the DPM-Solver++ inversion strategy to obtain an inversion latent from input image $x_0$ for the editing task. Additionally, this technique was not used in previously UNet-based image editing methods. Furthermore, we observe that we can still obtain a good inversion latent map without using DDIM inversion to calculate the values of $\boldsymbol{\hat{y}}$.

\textbf{The unified control of attention mechanism.}
In the previous work Prompt to Prompt (P2P)~\cite{hertz2022prompt}, researchers demonstrate that the cross attention layers contain rich semantic information from prompt texts. This finding can edit images through replacing the cross attention maps between the source image and target image during the diffusion process. Specifically, the two commonly used text-guided cross attention strategies are cross attention replacement and cross-attention refinement. These two methods ensure the seamless flow of information from the target prompt to the source prompt, thereby guiding the latent map towards the desired direction.

Different from the cross attention, the self-attention mechanism in the diffusion transformer is utilized to guide the formation of image layout, a feature that cannot be accomplished by the cross-attention mechanism. As shown in Figure~\ref{fig:attn_map}, the object and layout information from the prompt are not fully captured in the query vectors of the transformer’s shallow layers but are well-represented in the deeper layers. Moreover, with an increasing number of transformer layers, the query vectors' ability to capture object details becomes clearer and more specific. This suggests that the transformer's global attention mechanism is more effective at capturing long-range object information, making DiT particularly advantageous for large-scale deformation and editing of extensive images. This observation suggests that non-rigid editing of images can be achieved by controlling self attention mechanism. In MasaCtrl~\cite{cao2023masactrl}, researchers introduced the mutual self attention control mechanism. To be specific, in the early steps of diffusion, the feature in the editing steps $Q_{tar}$, $K_{tar}$, and $V_{tar}$ will be used in self attention calculation to generate an image layout closer to the target prompt, while in the later stages, the feature in the reconstruction steps ––$K_{src}$ and $V_{src}$ will be used to guide the generation of the target image layout closer to the original image.

However, MasaCtrl may still encounter some failure cases, which can be caused by its use of $Q_{tar}$ throughout the entire editing process, as mentioned in a recent work~\cite{Xu_2024_CVPR}.
To address this issue, we determine when to adopt $Q_{src}$ by setting a threshold $S$ for the number of steps:

\begin{equation}
\text{Mutual Edit} = 
\begin{cases}
\text{Attention}\{Q_{\text{src}},K_{\text{src}}, V_{\text{src}}\}, \text{ if } t > S \\
\text{Attention}\{Q_{\text{tar}}, K_{\text{src}}, V_{\text{src}}\}, \text{ otherwise}
\end{cases}
\label{equation: mutual attention}
\end{equation}

\begin{figure}[htp]
    \centering
    \includegraphics[width=8.5cm]{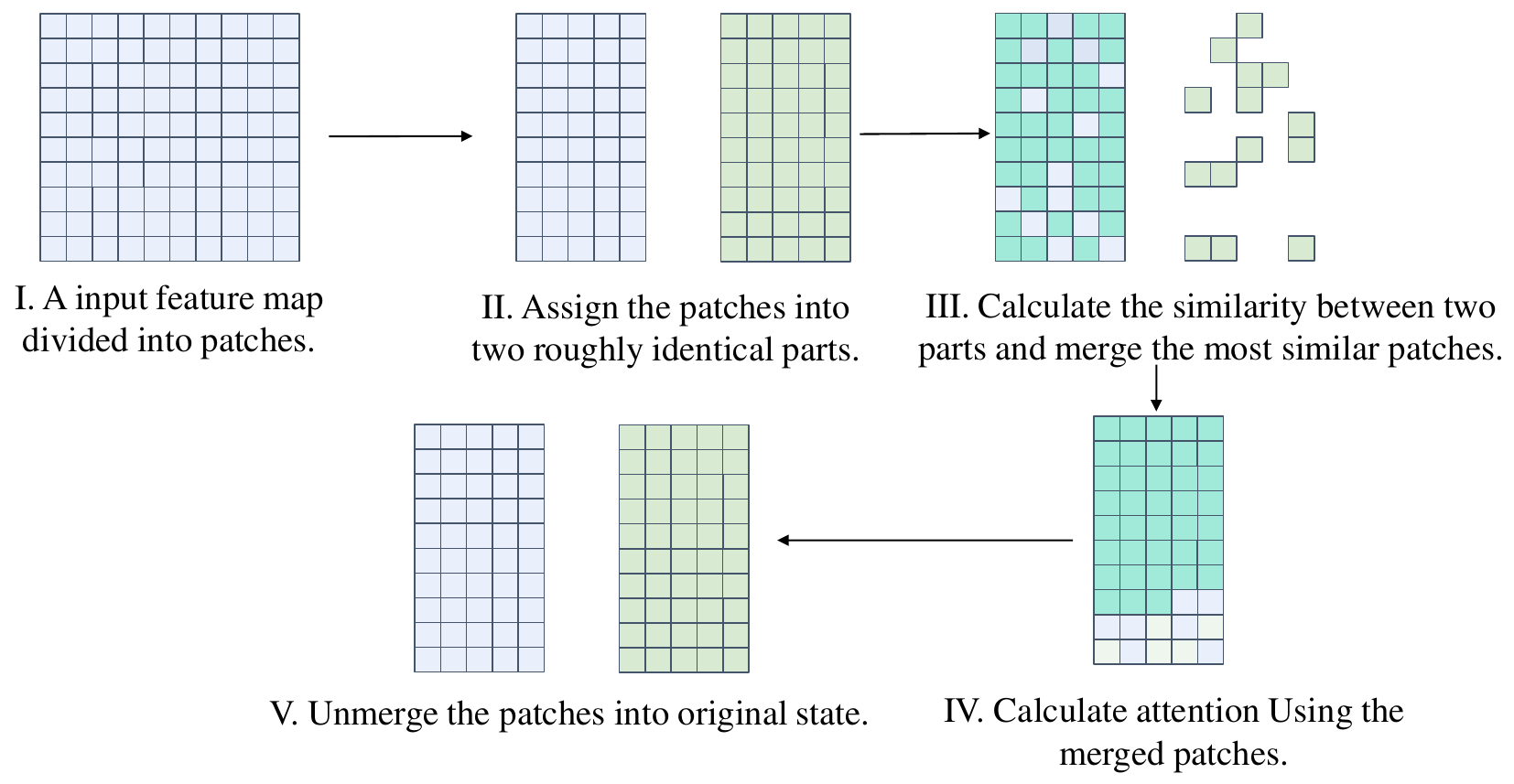}
    \caption{The calculation of patches merging.}
    \label{fig:patch_merging}
    
\end{figure}
\textbf{Patches merging.} To enhance the inference speed, inspired by token merging~\cite{bolya2023token}, we embed patches merging into the denoising model. This approach is motivated by the observation that the number of patches involved in attention calculations within the transformer architecture is significantly greater than that in UNet. The calculation flow is shown in Figure~\ref{fig:patch_merging}. For a feature map, we first compute the similarity between each patch and merge the most similar ones to reduce the number of patches processed by the attention mechanism. After attention calculation, we unmerge the patches to maintain the original input size for the next layer in the model. By incorporating patches merging into our framework, we aim to streamline the process and improve overall efficiency, without altering the fundamental operations of each layer. 
\begin{figure*}[t!]
    \centering
    \includegraphics[width=.98\textwidth]{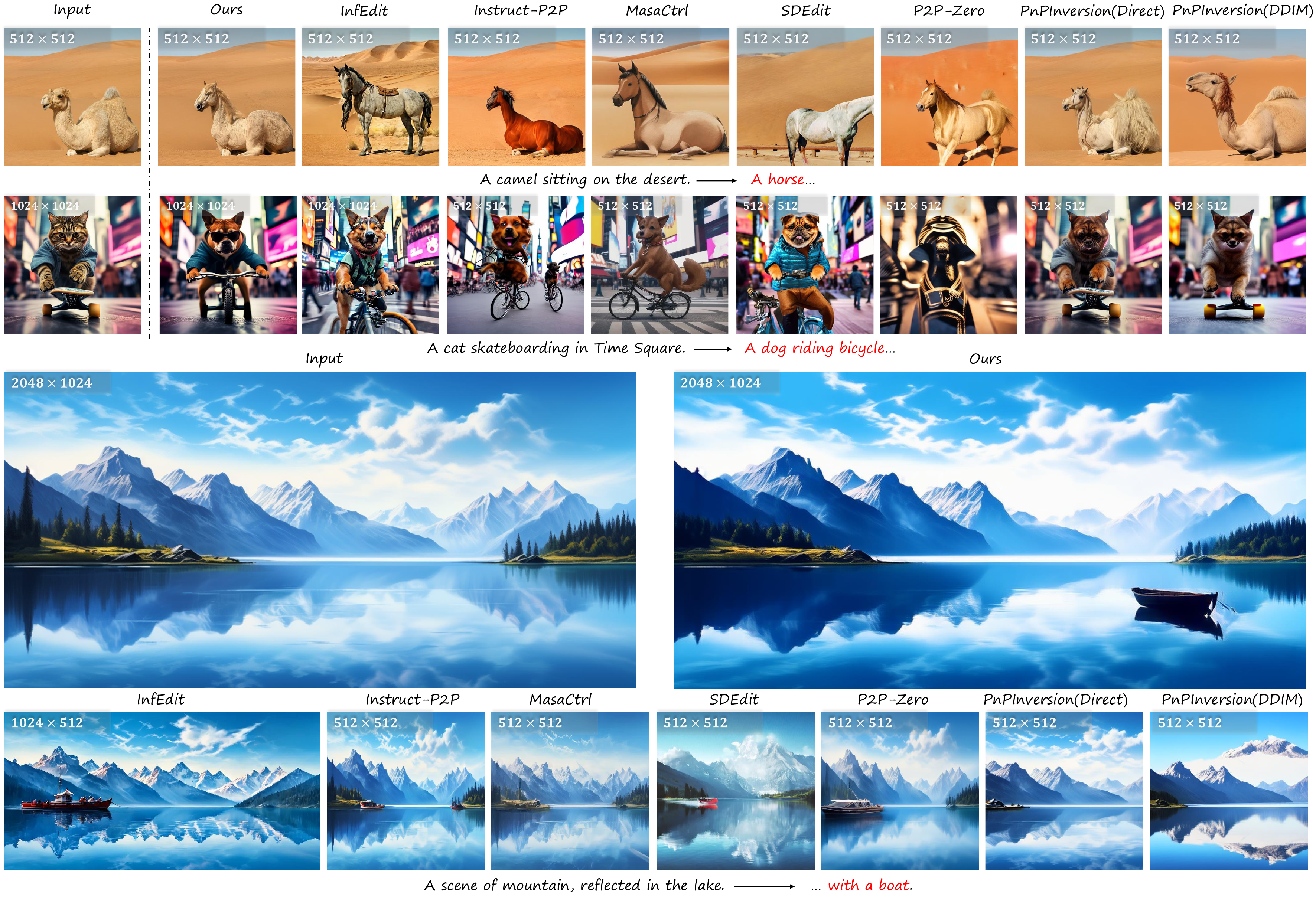}
    \vspace{-0.2cm}
    \caption{Comparative experiment with $512\times512$, $1024\times1024$, and high-resolution, non-typical aspect-ratio images against the baseline. DiT4Edit achieves satisfactory consistency in editing results.
    } 
    \label{fig:comparison-512}
\end{figure*}

\begin{table*}[t!]
\centering
\label{my-table}
\resizebox{1.0\textwidth}{!}{
\begin{tabular}{c|c|ccc|ccc|ccc|c}
\toprule
Metrics & &\multicolumn{3}{c|}{FID}  & \multicolumn{3}{c|}{PSNR} & \multicolumn{3}{c|}{CLIP} & Inference Time (s) \\
\cmidrule(lr){1-12}
\multicolumn{1}{c|}{Models}&\multicolumn{1}{c|}{Structure}&    \multicolumn{1}{c}{$512\times512$} & \multicolumn{1}{c}{$1024\times1024$} &  \multicolumn{1}{c|}{$1024\times2048$} & \multicolumn{1}{c}{$512\times512$} & \multicolumn{1}{c}{$1024\times1024$} &  \multicolumn{1}{c|}{$1024\times2048$} & \multicolumn{1}{c}{$512\times512$} & \multicolumn{1}{c}{$1024\times1024$} &  \multicolumn{1}{c|}{$1024\times2048$} & $512\times512$\\
\cmidrule(lr){1-12}
SDEdit~\cite{meng2021sdedit} &\multirow{6}{*}{UNet-based}&93.25&88.56&143.87&21.54&20.76&16.35&22.41&21.39&20.73&15.62\\ 
IP2P~\cite{brooks2023instructpix2pix}& &88.63&98.73&103.52&20.36&19.73&17.13&20.64&19.36&21.53&12.43\\
Pix2Pix-Zero~\cite{parmar2023zero}& &92.31&101.32&158.29&20.6&16.27&23.58&22.12&17.85&20.39&31.45\\
MasaCtrl~\cite{cao2023masactrl}& &110.75&176.15&236.49&17.35&20.51&16.92&23.51&21.96&15.23&19.76\\
InfEdit~\cite{xu2023inversion}& &86.28&87.42&98.75&21.54&22.36&21.74&24.46&23.74&22.16&\textbf{5.08}\\
PnPInversion~\cite{ju2024pnp}&  &84.73&85.33&110.73&21.71&23.42&24.59&21.71&20.76&19.35&30.48\\
\cmidrule(lr){1-12}
Ours&DiT-basd&\textbf{72.36}&\textbf{62.45}&\textbf{75.43}&\textbf{22.85}&\textbf{29.75}&\textbf{27.46}&\textbf{25.39}&\textbf{26.97}&\textbf{25.66}&5.15\\
\bottomrule
\end{tabular}
}
\caption{Quantitative comparison results. We compare our model with six prior works, all implemented using official open source code.}
\label{tab:quantitative}
\end{table*}

\section{Experiments}
\subsection{Implementation Details}
For editing tasks involving images with a scale of $512\times512$ and larger sizes up to $1024\times2048$, we use pre-trained models PixArt-$\alpha-\text{XL}-512\times512$ version for the smaller scale and PixArt-$\alpha-\text{XL}-1024\times1024-\text{MS}$ version for the larger scale~\cite{chen2023pixart}. We conduct editing on both real and generated images. For the real image input, we use DPM-Solver inversion to get the latent noise map. We configured the DPM-Solver with 30 steps, the classifier-free guidance of 4.5, and a patches merging ratio of 0.8. All experiments were carried out using an NVIDIA Tesla A100 GPU.

\subsection{Qualitative Comparison}
We evaluate the qualitative performance differences between our proposed DiT4Edit editing framework and six prior baselines, all implemented using official open source code.

As shown in Figure~\ref{fig:comparison-512}, we compare our method on $512\times512$ and $1024\times1024$ images. The first row of the Figure~\ref{fig:comparison-512} demonstrates our framework has the ability to generate edited images that remain consistent with the original content when editing real $512\times512$ images, whereas existing methods often alter the background or target details of the original image. Furthermore, the second and third rows of Figure~\ref{fig:comparison-512} illustrate our experiments with large-scale images and arbitrarily sized images—tasks that previous UNet-based methods struggled to address. The results indicate that our proposed framework effectively handles style and object shape modifications in larger images. In contrast, some state-of-the-art UNet-based methods, despite being capable of performing editing tasks, frequently result in significant alterations and damage to the background and object locations in the source image. Additionally, due to the limitations of the UNet structure, these methods typically generate target images only at a size of $512\times512$. These findings emphasize the substantial potential of transformer-based diffusion models in large-scale image editing. We also perform the user study for comprehensive comparisons. The details of the user study can be found in the supplementary material.


\subsection{Quantitative Comparison}
For quantitative evaluation, we used three indicators: Fréchet Inception Distance (FID)~\cite{NIPS2017_8a1d6947}, Peak Signal-to-Noise Ratio (PSNR), and CLIP to evaluate the performance differences between our model and SOTA in image generation quality, background preservation, and text alignment. We compared images at three sizes: $512\times512$, $1024\times1024$, and $1024\times2048$, with results detailed in Table~\ref{tab:quantitative}. We perform the performances with Pix2Pix-Zero, PnPInversion, SDEdit, IP2P, MasaCtrl, and InfEdit. It should be noted that, since no DiT-based editing framework previously existed, all our comparison baselines are based on the UNet architecture. The experimental results show that our proposed DiT4Edit editing strategy outperforms SOTA methods in image generation quality, background preservation, and text alignment. Due to the global attention capabilities of the integrated transformer structure, the DiT4Edit framework exhibits strong robustness across editing tasks of various sizes. The generated images not only show higher quality but also offer better control over the background and details, resulting in greater consistency with the original image. Particularly for editing large or arbitrarily sized images, DiT4Edit demonstrates significant advantages over other methods, showcasing the powerful scaling ability of the transformer architecture. Meanwhile, our editing framework has a shorter inference time, comparable to the inversion free editing method (InfEdit).

\subsection{Ablation Study}
We perform a series of ablation studies to demonstrate the effectiveness of \textbf{DPM-Solver inversion} and \textbf{patches merging}. The results of our ablation experiments on \textbf{patches merging} are presented in Figure~\ref{fig:patch_merging} and Table~\ref{tab:ablation_patch_merging}. Implementing patches merging led to a notable reduction in the editing time for large-sized images while maintaining editing quality comparable to that achieved without patches merging. This indicates that patches merging can significantly enhance the overall performance of image editing frameworks.
Furthermore, the ablation experiment results for \textbf{DPM-Solver and DDIM} are illustrated in Figure~\ref{fig:ablation_dpm}. When comparing the two methods with the same number of inference steps ($T=30$), DPM Solver consistently outperformed DDIM in terms of image editing quality. This demonstrates that our use of the DPM-Solver inversion strategy allows for the generation of superior latent maps, resulting in better editing outcomes within fewer steps.

\begin{table}[h!]
    \centering
    \resizebox{0.6\columnwidth}{!}{
    \begin{tabular}{ccc}
        \toprule
        \textbf{Image Size} & \textbf{Patches Merging} & \textbf{Speed (s)}\\
        \midrule
        \multirow{2}*{$512\times512$} &\ \Checkmark & 6.01\\
        
        & \XSolidBrush &9.13 \\
        
        \multirow{2}*{$1024\times1024$} &\ \Checkmark & 34.27\\
        
        & \XSolidBrush &41.52 \\
        \multirow{2}*{$1024\times2048$} &\ \Checkmark & 99.39\\
        
        & \XSolidBrush &122.41 \\
        \bottomrule

    \end{tabular}}
        \caption{Ablation study on patches merging. The result demonstrates that this technique accelerates model inference speed, especially for large-sized image editing, without affecting the quality of the final image generation.}
    \label{tab:ablation_patch_merging}
\end{table}

\begin{figure}[t!]
    \centering
    \includegraphics[width=0.99\columnwidth]{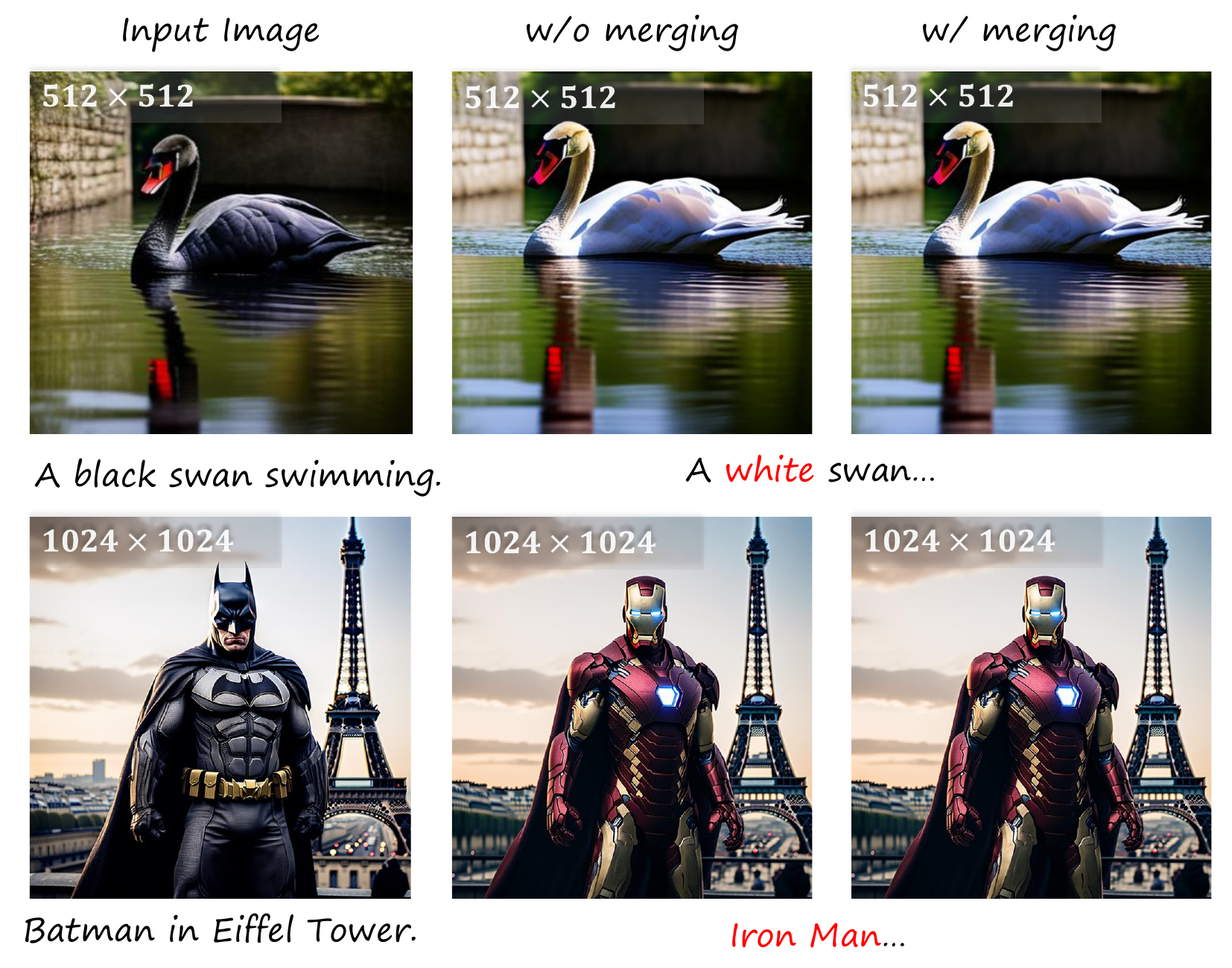}
    \vspace{-0.02cm}
    \caption{Ablation study on patches merging. The results indicate that this module does not impact the final quality of image editing.}
    \label{fig:ablation_patch_merging}
\end{figure}

\begin{figure}[t!]
    \centering
    \includegraphics[width=0.99\columnwidth]{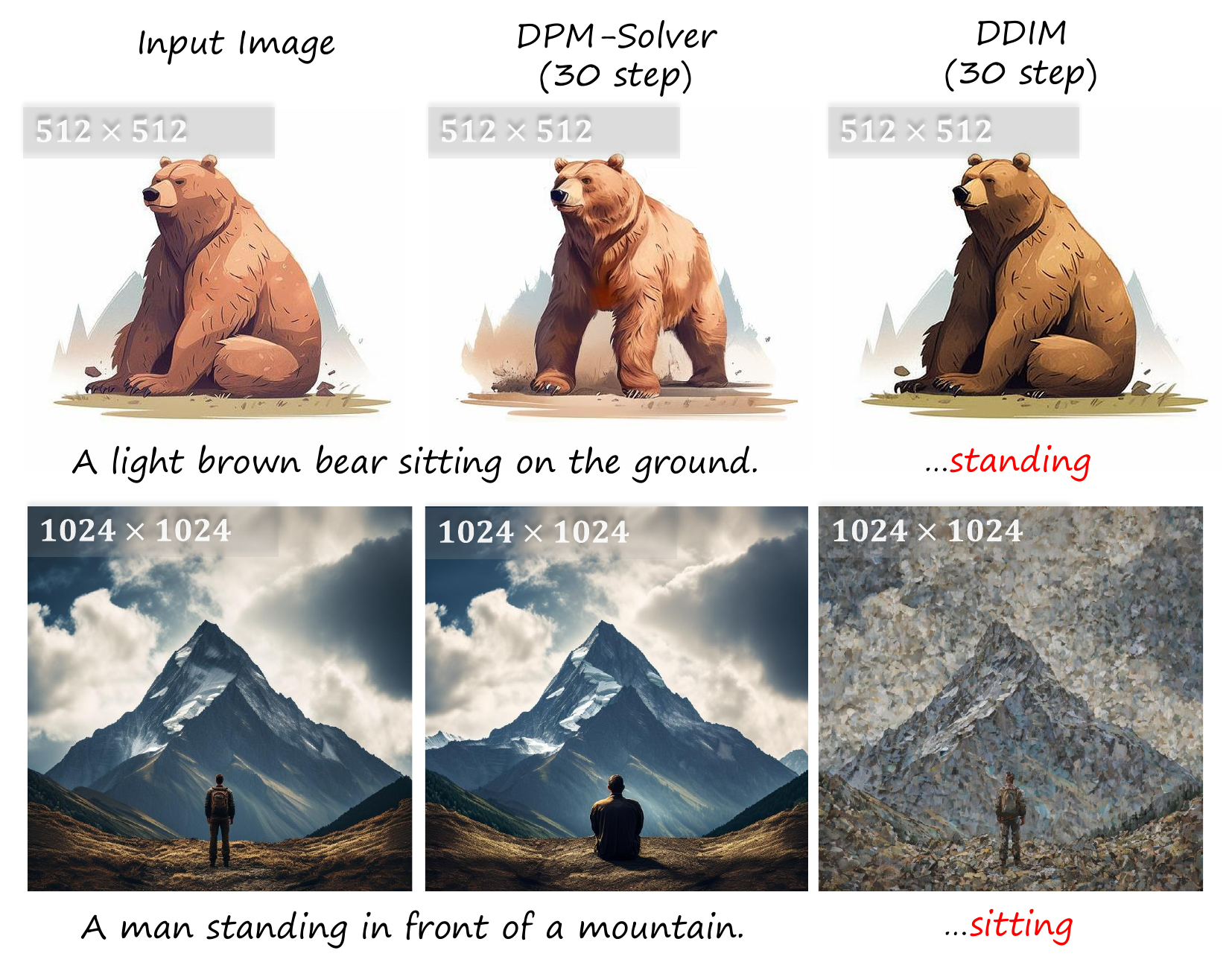}
    \vspace{-0.02cm}
    \caption{Ablation study on DPM-Solver inversion. Experiments have demonstrated that the DPM solver achieves superior editing results compared to DDIM, with fewer inference steps.}
    \label{fig:ablation_dpm}
\end{figure}

\section{Discussion and Conclusion}
\textbf{Conclusion}.
We introduce DiT4Edit, the first image-editing framework based on a diffusion transformer. Unlike previous UNet-based frameworks, DiT4Edit offers superior editing quality and supports images of various sizes. Leveraging DPM Solver inversion, a unified attention control mechanism, and patch merging, DiT4Edit outperforms the UNet structure in editing tasks for images sized $512\times512$ and $1024\times1024$. Notably, DiT4Edit can handle images of arbitrary sizes, such as $1024\times2048$, showcasing the transformer's advantages in global attention and scalability. Our research can set a baseline for DiT-based image editing and help further explore the potential of transformer structures in generative AI.
\\\textbf{Limitation}.
In our experiment, we observed that the T5-tokenizer occasionally encounters issues with word segmentation, which can lead to failures in the final editing process. Additionally, our model might experience color inconsistencies compared to the original image. Further editing failures are provided in the supplementary materials.
\\\textbf{Potential social impact}.
Advances in image editing models open doors for artistic innovation but also present risks. These include challenges in assessing image authenticity and potential privacy breaches from unauthorized edits. Clear standards and regulations are needed to ensure responsible use and mitigate these risks. Future model development will prioritize addressing these concerns.


\end{document}